# LEARNING NONSINGULAR PHYLOGENIES AND HIDDEN MARKOV MODELS

By Elchanan Mossel[1] and Sébastien Roch[2]

*University of California, Berkeley*

In this paper we study the problem of learning phylogenies and hidden Markov models. We call a Markov model nonsingular if all transition matrices have determinants bounded away from 0 (and 1). We highlight the role of the nonsingularity condition for the learning problem. Learning hidden Markov models without the nonsingularity condition is at least as hard as learning parity with noise, a well-known learning problem conjectured to be computationally hard. On the other hand, we give a polynomial-time algorithm for learning nonsingular phylogenies and hidden Markov models.

**1. Introduction.** In this paper we consider the problem of learning phylogenies and hidden Markov models, two of the most popular Markov models used in applications.

Phylogenies are used in evolutionary biology to model the stochastic evolution of genetic data on the ancestral tree relating a group of species. More precisely, the leaves of the tree correspond to (known) extant species. Internal nodes represent extinct species, while the root of the tree represents the most recent ancestor to all species in the tree. Following paths from the root to the leaves, each bifurcation indicates a speciation event whereby two new species are created from a parent.

The underlying assumption is that genetic information evolves from the root to the leaves according to a Markov model on the tree. This genetic information may consist of DNA sequences, proteins, and so on. Suppose,

Received June 2005; revised October 2005.
[1]Supported in part by a Miller Fellowship in Statistics and Computer Science, U.C. Berkeley, by a Sloan Fellowship in Mathematics and by NSF Grants DMS-05-04245 and DNS-05-28488.
[2]Supported by CIPRES (NSF ITR Grant NSF EF 03-31494), FQRNT, NSERC and a Loève Fellowship.

*AMS 2000 subject classifications.* 60J10, 60J20, 68T05, 92B10.

*Key words and phrases.* Hidden Markov models, evolutionary trees, phylogenetic reconstruction, PAC learning.

for example, that the genetic data consists of (aligned) DNA sequences and consider the evolution of the first letter in all sequences. This collection, named the first *character*, evolves according to Markov transition matrices on the edges. The root is assigned one of the four letters $A, C, G$ and $T$. Then this letter evolves from parents to descendants according to the Markov matrices on the edges connecting them.

The map from each node of the tree to the $i$th letter of the corresponding sequence is called the $i$th *character*. It is further assumed that the characters are i.i.d. random variables. In other words, each site in a DNA sequence is assumed to mutate independently from its neighbors according to the same mutation mechanism. Naturally, this is an over-simplification of the underlying biology. Nonetheless, the model above may be a good model for the evolution of some DNA subsequences and is the most popular evolution model in molecular biology, see, for example, [12].

One of the major tasks in molecular biology, the *reconstruction of phylogenetic trees*, is to infer the topology of the (unknown) tree from the characters (sequences) at the leaves (extant species). Often one is also interested in inferring the Markov matrices on the edges. In this paper we pay special attention to computational issues arising from the problem of inferring the complete Markov model. We seek to design *efficient* reconstruction algorithms or provide evidence that such algorithms do not exist. Here, efficiency means both that the length of the sequences used is polynomial in the number of leaves—that is, information-theoretic efficiency—and that the number of elementary steps performed by the reconstruction algorithm is polynomial in the number of leaves—that is, complexity-theoretic efficiency. The main technique for proving that a computational problem does not admit an efficient algorithm is to show that a well-known hard problem is a special case of this new problem. This is called a reduction, the most common of which is an NP-hardness reduction [16]. Let $n$ be the number of leaves; then the fact that a sequence of length $\Omega(n^c)$, with $c > 0$, is needed is established in [25].

In the systematics and statistics literature, three main approaches have been studied in depth for the reconstruction problem: parsimony, maximum likelihood and distance-based methods (see, e.g., [12, 30] for a detailed review and a thorough bibliography). Parsimony is known to be inconsistent [15] (it may converge to the wrong tree even if the number of characters tends to infinity) and NP-hard [18]. Maximum likelihood is also NP-hard [6, 29], but it is consistent [5]. As for distance-based methods, they can be consistent and, furthermore, run in polynomial time [10] (under some assumptions, see below). However, these methods have not gained popularity in biology yet (see [28]).

Much work has been devoted to the reconstruction of phylogenies in the learning setting [2, 7, 13]. In particular, in [7] a polynomial-time algorithm is obtained for Markov models on 2-state spaces in complete generality (in particular, there are no assumptions of regularity of the Markov transition matrices). Note that the authors of [7] also conjecture that their technique extends to the general $k$-state model, but then restrict themselves to $k = 2$.

One may roughly summarize the different approaches to phylogenetic reconstruction as follows:

- In biology the interest is usually in reconstructing the tree topology, as well as the full Markov model. However, most work in biology deals with special reconstruction methods and there are no results on efficient reconstruction.
- Work in combinatorial phylogeny focuses on efficient reconstruction of the tree topology. Here the Markov mutation model on the tree is not reconstructed.
- Work in learning theory shows that tree topologies and Markov models can be efficiently reconstructed when the number of character states is 2. In this paper we discuss the problem of recovering trees and Markov models when the number of character states is more than 2.

The framework of Markov models on trees has several special cases that are of independent interest. The case of mixtures of product distributions is discussed in [14]. Arguably, the most interesting special case is that of learning hidden Markov models (HMMs). HMMs play a crucial role in many areas from speech recognition to biology; see, for example, [8, 27]. It is easy to encode an HMM as a Markov model on a caterpillar tree. See Figure 1, where the states on the top line correspond to the hidden states and the states at the bottom correspond to observed states. The arrows going downward correspond to functions applied to the hidden states.

In [1] it is shown that finding the "optimal" HMM for an arbitrary distribution is hard unless $RP = NP$ (it is widely believed that $RP \neq NP$). See also [24] where hardness of approximation results are obtained for problems such as comparing two hidden Markov models. Most relevant to our setting is the conjecture made in [22] that learning parity with noise is computationally hard. It is easy to see that the problem of learning parity with noise may be encoded as learning an HMM over 4 states. See Section 1.3.

There is an interesting discrepancy between the two viewpoints taken in works concerning learning phylogenies and works concerning learning hidden Markov models. The results in phylogeny are mostly positive—they give polynomial-time algorithms for learning. On the other hand, the results concerning HMMs are mostly negative.

This paper tries to resolve the discrepancy between the two points of view by pointing to the source of hardness in the learning problem. Roughly speaking, it seems like the source of hardness for learning phylogenies and hidden Markov models are transition matrices $P$ such that $\det P$ is 0 (or close to 0) but $\mathrm{rank}\,P > 1$ (or $P$ is far from a rank 1 matrix). Note that, in the case $k = 2$, there are no matrices whose determinant is 0 and whose rank is more than 1. Indeed, in this case, the problem is not hard [7]. We note, furthermore, that in the problem of learning parity with noise all of the determinants are 0 and all the ranks are greater than 1.

The main technical contribution of this paper is to show that the learning problem is feasible once all the matrices have $\beta < |\det P| < 1 - 1/\mathrm{poly}(n)$ for some $\beta > 0$. We thus present a proper PAC learning algorithm for this case. In the case of hidden Markov models we prove that the model can be learned under the weaker condition that $1/\mathrm{poly}(n) < |\det P| \leq 1$. Assuming that learning parity with noise is indeed hard, this is an optimal result. See Section 1.3.

The learning algorithms we present are based on a combination of techniques from phylogeny, statistics, combinatorics and probability. We believe that these algorithms may also be extended to cases where $|\det P|$ is close to 1 and, furthermore, to cases where if $|\det P|$ is small, then the matrix $P$ is close to a rank 1 matrix, thus, recovering the results of [7].

Interestingly, to prove our result, we use and extend several previous results from combinatorial phylogeny and statistics. The topology of the tree is learned via variants of combinatorial results proved in phylogeny [10]. Thus, the main technical challenge is to learn the mutation matrices along the edges. For this, we follow and extend the approach developed in statistics by Chang [5]. Chang's results allow the recovery of the mutation matrices from an infinite number of samples. The reconstruction of the mutation matrices from a polynomial number of samples requires a delicate error analysis along with various combinatorial and algorithmic ideas.

The algorithm is sketched in Section 2 and the error analysis is detailed in Section 3.

1.1. *Definitions and results.* We let $\mathbf{T}_3(n)$ denote the set of all trees on $n$ labeled leaves where all internal degrees are exactly 3. Note that if $\mathcal{T} = (\mathcal{V}, \mathcal{E}) \in \mathbf{T}_3(n)$, then $|\mathcal{V}| = 2n - 2$. We will sometimes omit $n$ from the notation. Below we will always assume that the leaf set is labeled by the set $[n] = \{1, \ldots, n\}$. We also denote the leaf set by $\mathcal{L}$. Two trees $\mathcal{T}_1, \mathcal{T}_2$ are considered identical if there is a graph isomorphism between them that is the identity map on the set of leaves $[n]$. We define a *caterpillar* to be a tree on $n$ leaves with the following property: the subtree induced by the internal nodes is a path (and all internal vertices have degree at least 3). See Figure 1 for an example. We let $\mathbf{TC}_3(n)$ denote the set of all caterpillars on $n$ labeled leaves.

In a Markov model $T$ on a (undirected) tree $\mathcal{T} = (\mathcal{V}, \mathcal{E})$ rooted at $r$, each vertex iteratively chooses its state given the one of its parent by an application of a Markov transition rule. Consider the orientation of $\mathcal{E}$ where all edges are directed away from the root. We note this set of directed edges $\mathcal{E}_r$. Then the probability distribution on the tree is the probability distribution on $\mathcal{C}^\mathcal{V}$ given by

$$\pi^T(s) = \pi_r^T(s(r)) \prod_{(u,v)\in\mathcal{E}_r} P^{uv}_{s(u)s(v)}, \tag{1}$$

where $s \in \mathcal{C}^\mathcal{V}$, $\mathcal{C}$ is a finite state space, $P^{uv}$ is the transition matrix for edge $(u,v) \in \mathcal{E}_r$ and $\pi_r^T$ is the distribution at the root. We let $k = |\mathcal{C}|$. We write $\pi_\mathcal{W}^T$ for the marginals of $\pi^T$ on the set $\mathcal{W}$. Since the set of leaves is labeled by $[n]$, the marginal $\pi_{[n]}^T$ is the marginal on the set of leaves. We will often remove the superscript $T$. Furthermore, for two vertices $u, v \in \mathcal{V}$, we let $P^{uv}_{ij} = \mathbb{P}[s(v) = j | s(u) = i]$. We will be mostly interested in *nonsingular* Markov models.

DEFINITION 1. We say that a Markov model on a tree $\mathcal{T} = (\mathcal{V}, \mathcal{E})$ is *nonsingular* $[(\beta, \beta', \sigma)$-*nonsingular*$]$ if we have the following:

I. For all $e \in \mathcal{E}_r$, it holds that $1 > |\det P^e| > 0$ $[1 - \beta' \geq |\det P^e| > \beta]$ and
II. For all $v \in \mathcal{V}$, it holds that $\pi_v(i) > 0$ $[\pi_v(i) > \sigma]$ for all $i$ in $\mathcal{C}$.

It is well known [31] that if the model is nonsingular, then, for each $w \in \mathcal{V}$, one can write

$$\pi^T(s) = \pi_w^T(s(w)) \prod_{(u,v)\in\mathcal{E}_w} P^{uv}_{s(u)s(v)}, \tag{2}$$

where now all edges $(u, v)$ are oriented away from $w$. In other words, the tree may be rooted arbitrarily. Indeed, in the learning algorithms discussed below, we will root the tree arbitrarily. We will actually refer to $\mathcal{E}$ as the set of directed edges formed by taking the two orientations of all (undirected) edges in the tree. It is easy to show that $(\beta, \beta', \sigma)$-nonsingularity as stated above also implies that property I holds for all $(u, v) \in \mathcal{E}$ with appropriate values of $\beta, \sigma$.

Transition matrices $P$ satisfying $|\det P| = 1$ are permutation matrices. While edges equipped with such matrices preserve information, it is impossible to deduce the existence of such edges from the phylogenetic data. For example, if all edges satisfy that $P$ is the identity matrix, then the characters are always constant for all possible trees.

Note, moreover, that if $|\det P^{uv}| > 0$ for all edges $(u, v)$ and for all $v \in \mathcal{V}$, the distribution of $s(v)$ is supported on at most $|\mathcal{C}| - 1$ elements, then one can redefine the model by allowing only $|\mathcal{C}| - 1$ values of $s(v)$ at each node and deleting the corresponding rows and columns from the transition matrices $P^e$. (Note that the labels of the character states at internal nodes are in fact determined only up to a permutation and similarly for the mutation matrices. This will be explained in more detail below.) Thus, condition II is very natural given condition I.

We call a Markov model as in (1) a phylogenetic tree. Given collections $\mathbf{M}_n$ of mutation matrices $P$ and collections $\mathbf{T}(n)$ of trees on $n$ leaves, we let $\mathbf{T}(n) \otimes \mathbf{M}_n$ denote all phylogenetic trees of the form (1), where $\mathcal{T} \in \mathbf{T}(n)$ and $P^e \in \mathbf{M}_n$ for all $e$. Given numbers $0 \leq \sigma_n < 1$, we write $(\mathbf{T}(n) \otimes \mathbf{M}_n, \sigma_n)$ for all the elements of $\mathbf{T}_3(n) \otimes \mathbf{M}_n$ satisfying $\pi_v(i) > \sigma_n$ for all $i \in \mathcal{C}$ and $v \in \mathcal{V}$. We will be particularly interested in the collections of all binary trees on $n$ leaves denoted $\mathbf{T}_3(n)$ and in the collections of all caterpillars on $n$ leaves denoted $\mathbf{TC}_3(n)$. Our goal is to provide efficient algorithms to infer the models $\mathbf{T}_3(n) \otimes \mathbf{M}_n$ and $\mathbf{TC}_3(n) \otimes \mathbf{M}_n$, given independent samples of the characters at the leaves. However, given any finite amount of data, one cannot hope to estimate exactly with probability 1 the tree and the transition matrices. Moreover, some degrees of freedom, such as the labels of the characters at the internal nodes, cannot be recovered even from the exact distribution at the leaves.

Since the model cannot be recovered exactly, an alternative approach is needed. The standard approach in computational learning theory is to use the PAC learning framework introduced by [32], here in its variant proposed by [22]. PAC learning has been studied extensively in the learning theory literature. For background and references, see [23].

Let $\varepsilon > 0$ denote an approximation parameter, $\delta > 0$, a confidence parameter, $(\mathbf{M}_n)$, collections of matrices, $(\mathbf{T}(n))$, collections of trees, and $(\sigma_n)$, a sequence of positive numbers. Then we say that an algorithm $\mathcal{A}$ *PAC-learns* $(\mathbf{T}(n) \otimes \mathbf{M}_n, \sigma_n)$ if for all $n$ and all $T \in (\mathbf{T}(n) \otimes \mathbf{M}_n, \sigma_n)$, given access to independent samples from the measure $\pi_{[n]}^T$, $\mathcal{A}$ outputs a phylogenetic tree $T'$ such that the total variation distance between $\pi_{[n]}^T$ and $\pi_{[n]}^{T'}$ is smaller than $\varepsilon$ with probability at least $1 - \delta$ and the running time of $\mathcal{A}$ is $\mathrm{poly}(n, 1/\delta, 1/\varepsilon)$.

In our main result we prove the following.

THEOREM 1. *For every constant $\beta, \kappa_\beta, \kappa_\pi > 0$ and every finite set $\mathcal{C}$, the collection of $(\beta, n^{-\kappa_\beta}, n^{-\kappa_\pi})$-nonsingular Markov phylogenetic models is PAC-learnable. More formally, let $\mathcal{C}$ be a finite set, $\beta, \kappa_\beta, \kappa_\pi > 0$. Let $\mathbf{M}_n$ denote the collection of all $|\mathcal{C}| \times |\mathcal{C}|$ transition matrices $P$, where $1 - n^{-\kappa_\beta} > |\det P| > \beta$. Then there exists a PAC-learning algorithm for $(\mathbf{T}_3(n) \otimes \mathbf{M}_n, n^{-\kappa_\pi})$ whose running time is $\mathrm{poly}(n, |\mathcal{C}|, 1/\varepsilon, 1/\delta)$.*

*Furthermore, if the learning problem has an additional input which is the true tree topology, then the assumption on determinants in $\mathbf{M}_n$ can be relaxed to $1 \geq |\det P| > \beta$.*

For hidden Markov models, we can prove more.

THEOREM 2. *Let $\phi_d, \kappa_\pi > 0$ be constants. Let $\mathcal{C}$ be a finite set and $\mathbf{M}_n$ denote the collection of $|\mathcal{C}| \times |\mathcal{C}|$ transition matrices $P$, where $1 \geq |\det P| > n^{-\phi_d}$. Then there exists a PAC-learning algorithm for $(\mathbf{TC}_3(n) \otimes \mathbf{M}_n, n^{-\kappa_\pi})$. The running time and sample complexity of the algorithm is $\mathrm{poly}(n, |\mathcal{C}|, 1/\varepsilon, 1/\delta)$.*

In many applications of HMMs, the state spaces at different vertices are of different sizes and, therefore, many of the Markov matrices have 0 determinant. Theorem 2 is not applicable in these cases. Indeed, then, the negative result presented in Section 1.3 may be more relevant.

1.2. *Inferring the topology.* We let the topology of $T$ denote the underlying tree $\mathcal{T} = (\mathcal{V}, \mathcal{E})$. The task at hand can be divided into two natural subproblems. First, the topology of $T$ needs to be recovered with high probability. Second, the transition matrices have to be estimated. Reconstructing the topology has been a major task in phylogeny. It follows from [10, 11] that the topology can be recovered with high probability using a polynomial number of samples. Here is one formulation from [26].

THEOREM 3. *Let $\beta > 0$, $\kappa_\beta > 0$ and suppose that $\mathbf{M}_n$ consists of all matrices $P$ satisfying $\beta < |\det P| < 1 - n^{-\kappa_\beta}$. For all $\kappa_T > 2$, the topology of $T \in (\mathbf{T}_3(n) \otimes \mathbf{M}_n, n^{-\kappa_\pi})$ can be recovered in polynomial time using $n^{O(1/\beta + \kappa_\beta + \kappa_T + \kappa_\pi)}$ samples with probability at least $1 - n^{2-\kappa_T}$.*

We will also need a stronger result that applies only to hidden Markov models. The proof, which is sketched in the Appendix, is quite similar to the proofs in [10, 11].

THEOREM 4. *Let $\zeta > 0, \kappa_\pi > 0$ and suppose that $\mathbf{M}_n$ consists of all matrices $P$ satisfying $n^{-\zeta} < |\det P| \leq 1$. Then for all $\theta > 0, \tau > 0$, and all $T \in (\mathbf{TC}_3(n) \otimes \mathbf{M}_n, n^{-\kappa_\pi})$, one can recover from $n^{O(\zeta + \theta + \tau + \kappa_\pi)}$ samples a tree-topology $\mathcal{T}'$ with probability $1 - n^{-\theta}$, where the topology $\mathcal{T}'$ satisfies the following. It is obtained from the true topology $\mathcal{T}$ by contracting some of the internal edges whose corresponding mutation matrices $P$ satisfy $|\det P| > 1 - n^{-\tau}$. Note that $\mathcal{T}'$ may have some of its internal degrees greater than 3.*

1.3. *Hardness of learning singular models.* We now briefly explain why hardness of learning "parity with noise" implies that learning singular hidden Markov models is hard. We first define the parity-learning problem, which has been extensively studied in the computational learning theory. See, in particular, [3, 4, 19, 21].

DEFINITION 2 (Learning parity with noise). Let $\mathbf{x} = (x_1, \ldots, x_n)$ be a vector in $\{0,1\}^n$, $T$ a subset of $\{1,\ldots,n\}$ and $0 < \alpha < 1/2$. The parity of $\mathbf{x}$ on $T$ is the Boolean function, denoted $\phi_T(\mathbf{x}) = \bigoplus_{i \in T} x_i$, which outputs 0 if the number of ones in the subvector $(x_i)_{i \in T}$ is even and 0 otherwise. A uniform query oracle for this problem is a randomized algorithm that returns a random uniform vector $\mathbf{x}$, as well as a noisy classification $f(\mathbf{x})$ which is equal to $\phi_T(\mathbf{x})$ with probability $\alpha > 0$ and $1 - \phi_T(\mathbf{x})$ otherwise. All examples returned by the oracle are independent. The learning parity with noise problem consists in designing an algorithm with access to the oracle such that, for all $\varepsilon, \delta > 0$, the algorithm returns a function $h : \{0,1\}^n \to \{0,1\}$ satisfying $\mathbb{P}_{\mathbf{x}}[h(\mathbf{x}) = \phi_T(\mathbf{x})] \geq 1 - \varepsilon$ (where $\mathbf{x}$ is uniform over $\{0,1\}^n$) with probability at least $1 - \delta$ in time polynomial in $n, 1/\varepsilon, 1/\delta$.

Kearns' work [21] on the statistical query model leads to the following conjecture.

CONJECTURE 1 (Noisy parity assumption [21]). There is an $\alpha$ with $0 < \alpha < 1/2$ such that there is no efficient algorithm for learning parity under uniform distribution in the PAC framework with classification noise $\alpha$.

In [22], this is used to show that learning probabilistic finite automata with an evaluator is hard. It is easy to see that the same construction works with the probabilistic finite automata replaced by an equivalent hidden Markov model (HMM) with 4 states (this is a special case of our evolutionary tree model when the tree is a caterpillar). The proof, which is briefly sketched in Figure 1, is left to the reader. We remark that all matrices in the construction have determinant 0 and rank 2. Note, moreover, that, by a standard coupling argument, it follows that if for all edges $(u,v)$ we replace the matrix $P^{uv}$ by the matrix $(1 - n^{-\tau})P^{uv} + n^{-\tau}I$, then the model given in Figure 1 and its variant induces undistinguishable distributions on $K$ samples if $K \leq o(n^{\tau-1})$. This shows that, assuming that learning parity with noise is hard, Theorem 2 is tight up to the constant in the power of $n$.

## 2. Overview of the algorithm.

2.1. *Chang's spectral technique.* One of the main ingredients of the algorithm is the following important result due to Chang [5] that we rederive here for completeness. Let $\mathcal{T}$ be a 4-node (star) tree with a root $r$ and 3 leaves $a, b, c$. (See Figure 2.) Let $P^{uv}$ be the transition matrix between vertices $u$ and $v$, that is, $P^{uv}_{ij} = \mathbb{P}[s(v) = j | s(u) = i]$ for all $i, j \in \mathcal{C}$. Fix $\gamma \in \mathcal{C}$. Then by the Markov property, for all $i, j \in \mathcal{C}$,

$$\mathbb{P}[s(c) = \gamma, s(b) = j | s(a) = i] = \sum_{h \in \mathcal{C}} P^{ar}_{ih} P^{rc}_{h\gamma} P^{rb}_{hj},$$

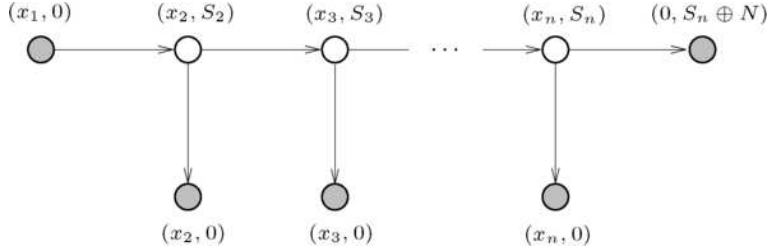

FIG. 1. *Hidden Markov model for noisy parity. The model computes $N \oplus \bigoplus_{i \in T} x_i$, where the $x_i$'s are uniform over $\{0,1\}$, $T$ is a subset of $\{1,\ldots,n\}$, and $N$ is a small random noise. The $S_i = \bigoplus_{j \in T, j \leq i} x_j$ are the partial sums over variables included in $T$. The observed nodes are in light gray. The hardness proof follows from a standard reduction technique similar to that used in* [22].

or, in matrix form, $P^{ab,\gamma} = P^{ar} \operatorname{diag}(P^{rc}_{\cdot\gamma}) P^{rb}$, where the matrix $P^{ab,\gamma}$ is defined by

$$P^{ab,\gamma}_{ij} = \mathbb{P}[s(c) = \gamma, s(b) = j | s(a) = i],$$

for all $i, j \in \mathcal{C}$. Then, noting that $(P^{ab})^{-1} = (P^{rb})^{-1}(P^{ar})^{-1}$, we have

$$(\star) \qquad P^{ab,\gamma}(P^{ab})^{-1} = P^{ar} \operatorname{diag}(P^{rc}_{\cdot\gamma})(P^{ar})^{-1},$$

assuming the transition matrices are invertible. Equation $(\star)$ is an eigenvalue decomposition where the l.h.s. involves only the distribution at the leaves. Therefore, given the distribution at the leaves, we can recover from $(\star)$ the columns of $P^{ar}$ (up to scaling), provided the eigenvalues are distinct. Note that the above reasoning applies when the edges $(r,a), (r,b), (r,c)$ are replaced by paths. Therefore, loosely speaking, in order to recover an edge $(w, w')$, one can use $(\star)$ on star subtrees with $w$ and $w'$ as roots to obtain $P^{aw}$ and $P^{aw'}$, and then compute $P^{ww'} = (P^{aw})^{-1} P^{aw'}$. In [5], under further assumptions on the structure of the transition matrices, the above scheme is used to prove the identifiability of the full model, that is, that the output distributions on triples of leaves uniquely determine the transition matrices. In this paper we show that the transition matrices can actually be approximately recovered using ($\star$) with a polynomial number of samples.

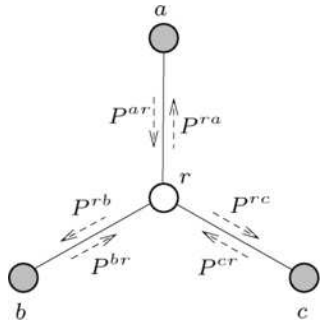

FIG. 2. *Star tree with three leaves.*

There are many challenges in extending Chang's identifiability result to our efficient reconstruction claim. First, as noted above, equation ($\star$) uncovers only the columns of $P^{ar}$. The leaves actually give no information on the labelings of the internal nodes. To resolve this issue, Chang assumes that the transition matrices come in a canonical form that allows to reconstruct them once the columns are known. For instance, if in each row, the largest entry is always the diagonal one, this can obviously be performed. This assumption is a strong and unnatural restriction on the model we wish to learn and, therefore, we seek to avoid it. The point is that relabeling all internal nodes does not affect the output distribution, and, therefore, the internal labelings can be made arbitrarily (in the PAC setting). The issue that arises is that those arbitrary labelings have to be made consistently over all edges sharing a node. Another major issue is that the leaf distributions are known only approximately through sampling. This requires a delicate error analysis and many tricks which are detailed in Section 3. The two previous problems are actually competing. Indeed, one way to solve the consistency issue is to fix a reference leaf $\omega$ and do all computations with respect to the reference leaf, that is, choose $a = \omega$ in every spectral decomposition. However, this will necessitate the use of long paths on which the error builds up exponentially. Our solution is to partition the tree into smaller subtrees, reconstruct consistently the subtrees using one of their leaves as a reference, and patch up the subtrees by fixing the connecting edges properly afterward. We refer to the connecting edges as *separators*.

A detailed version of the algorithm, FULLRECON, including two subroutines, appears in Figures 3, 4 and 5. The correctness of the algorithm uses the error analysis and is therefore left for Section 3. The two subroutines are described next.

2.2. *Subtree reconstruction and patching.* We need the following notation to describe the subroutines. If $e = (u,v)$, let $d_e(u)$ be the length of the shortest path (in number of edges) from $u$ to a leaf in $\mathcal{L}$ *not using edge $e$*. Then the depth of $\mathcal{T}$ is

$$\Delta = \max_{e=(u,v)\in\mathcal{E}} \{\max\{d_e(u), d_e(v)\}\}.$$

It is easy to argue that $\Delta = O(\log n)$. (See Section 3.) Also, for a set of vertices $\mathcal{W}$ and edges $\mathcal{S}$, denote $\mathcal{N}(\mathcal{W}, \mathcal{S})$ the set of nodes not in $\mathcal{W}$ that share an edge in $\mathcal{E}\setminus\mathcal{S}$ with a node in $\mathcal{W}$ ("outside neighbors" of $\mathcal{W}$ "without using edges in $\mathcal{S}$"). Let $\mathcal{B}_a^\Delta$ be the subset of nodes in $\mathcal{V}$ at distance at most $\Delta$ from leaf $a$.

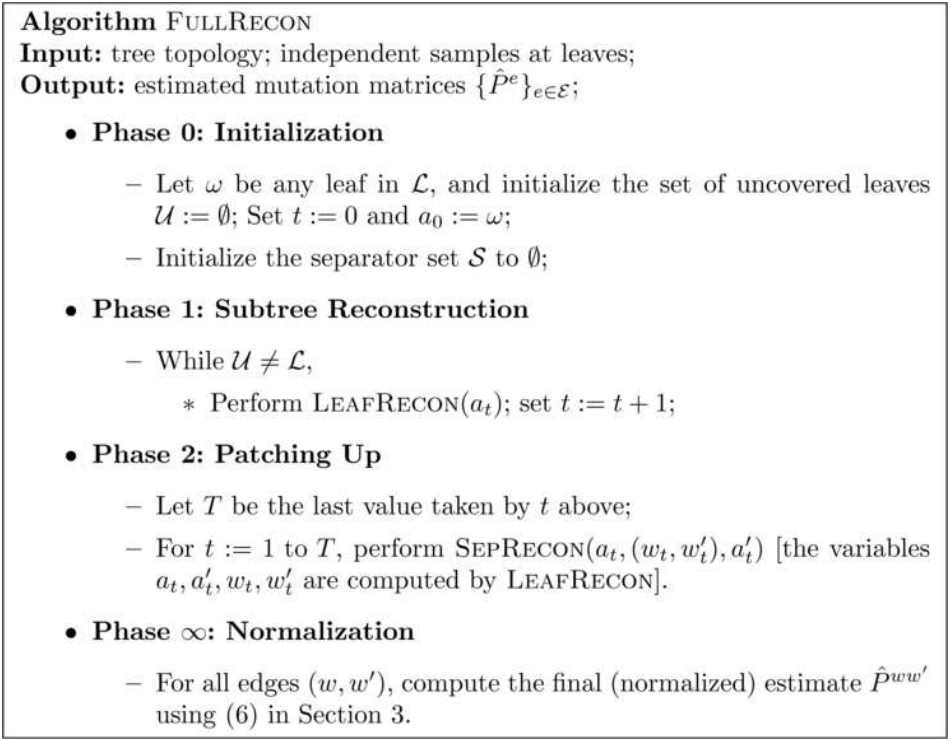

Fig. 3. *Algorithm* FULLRECON.

*Subroutine* LEAFRECON. The subtree reconstruction phase is performed by the algorithm LEAFRECON depicted in Figure 6. The purpose of the subroutine LEAFRECON is to partition the tree into subtrees, each of which has the property that all its nodes are at distance at most $\Delta$ from one of its leaves (same leaf for all nodes in the subtree), which we refer to as the reference leaf of the subtree. The correctness of the algorithm, proved in Section 3, thus establishes the existence of such a partition. This partition serves our purposes because it allows (1) to reconstruct mutation matrices in a consistent way (in each subtree) using reference leaves, and (2) to control the building up of the error by using short paths to the reference leaf. The matrix reconstruction is performed simultaneously by LEAFRECON, as the partition is built. At the call of LEAFRECON, we consider the subgraph $\mathcal{T}'$ of $\mathcal{T}$ where edges previously labeled as separators have been removed. We are given a reference leaf $a$ and restrict ourselves further to the (connected) subtree $\mathcal{T}_a$ of $\mathcal{T}'$ consisting of nodes at distance at most $\Delta$ from $a$. Moving away from $a$, we recover edge by edge the mutation matrices in $\mathcal{T}_a$ by Chang's spectral decomposition. At this point, it is crucial that (1) we use the transition matrices previously computed to ensure consistency in the

---

**Algorithm** LEAFRECON
**Input:** tree topology; reference leaf $a$; uncovered leaves $\mathcal{U}$; separator set $\mathcal{S}$; independent samples at the leaves;
**Output:** estimated mutation matrices $\tilde{P}^e$ and estimated node distributions $\tilde{\pi}_u$ on edges and vertices in subtree associated to $a$;

- Set $\mathcal{W} := \{a\}$, $\mathcal{U} := \mathcal{U} \cup \{a\}$;
- While $\mathcal{N}(\mathcal{W}, \mathcal{S}) \cap \mathcal{B}_a^\Delta \neq \emptyset$,
  - Pick any vertex $r \in \mathcal{N}(\mathcal{W}, \mathcal{S}) \cap \mathcal{B}_a^\Delta$;
  - Let $(r_0, r)$ be the edge with endpoint $r$ in the path from $a$ to $r$;
  - If $r_0 = a$, set $\tilde{P}^{ar_0}$ to be the identity, otherwise $\tilde{P}^{ar_0}$ is known from previous computations;
  - If $r$ is not a leaf,
    * Let $(r, r_1)$ and $(r, r_2)$ be the other two edges out of $r$;
    * Find the closest leaf $b$ (resp. $c$) connected to $r_1$ (resp. $r_2$) by a path not using $(r_0, r)$;
    * Draw Gaussian vector $U$; Perform the eigenvalue decomposition $(\star')$ (see Section 3);
    * Build $\hat{X}$ out of an arbitrary ordering of the columns recovered above;
    * Compute $\eta = (\hat{X})^{-1}$, and set $\tilde{P}^{ar} := \hat{X} \operatorname{diag}(\eta)$;
  - Otherwise if $r$ is a leaf:
    * Use estimate from leaves for $\hat{P}^{ar}$;
    * Set $\mathcal{U} := \mathcal{U} \cup \{r\}$;
  - Compute $\tilde{P}^{r_0 r} := (\tilde{P}^{ar_0})^{-1} \tilde{P}^{ar}$;
  - Compute $\tilde{\pi}_r = \hat{\pi}_a \tilde{P}^{ar}$;
  - Obtain $\tilde{P}^{rr_0}$ from $\tilde{P}^{r_0 r}$, $\tilde{\pi}_r$ and $\tilde{\pi}_{r_0}$ using Bayes rule;
  - Set $\mathcal{W} := \mathcal{W} \cup \{r\}$;
  - If $r$ is not a leaf and the distance between $a$ and $r$ is exactly $\Delta$, for $\iota = 1, 2$,
    * If $\{r, r_\iota\}$ is not in $\mathcal{S}$, set $t' := |\mathcal{S}| + 1$, $a_{t'} := b$ (if $\iota = 1$, and $c$ o.w.), $a'_{t'} := a$, $(w_{t'}, w'_{t'}) := (r_\iota, r)$, and add $\{r, r_\iota\}$ (as an undirected edge) to $\mathcal{S}$.

FIG. 4. *Algorithm* LEAFRECON.

labeling of internal nodes, and that (2) in order to control error we choose the leaves $b$ and $c$ (in the notation of the previous subsection) to be at distance at most $\Delta + 1$ from the edge currently reconstructed (which is always

**Algorithm** SEPRECON
**Input:** leaves $a$, $a'$ and separator edge $(w, w')$;
**Output:** estimated mutation matrices $\tilde{P}^{ww'}$, $\tilde{P}^{ww'}$;

- Estimate $\hat{P}^{aa'}$;
- Compute $\tilde{P}^{ww'} := (\tilde{P}^{aw})^{-1}\hat{P}^{aa'}(\tilde{P}^{w'a'})^{-1}$;
- Obtain $\tilde{P}^{w'w}$ from $\tilde{P}^{ww'}$, $\tilde{\pi}_w$ and $\tilde{\pi}_{w'}$ using Bayes rule.

FIG. 5. *Algorithm* SEPRECON.

possible by definition of $\Delta$). Note that the paths from the current node to $b$ and $c$ need not be in $\mathcal{T}_a$. Once $\mathcal{T}_a$ is reconstructed, edges on the "outside boundary" of $\mathcal{T}_a$ (edges in $\mathcal{T}'$ with exactly one endpoint in $\mathcal{T}_a$) are added to the list of separators, each with a new reference leaf taken from the unexplored part of the tree (at distance at most $\Delta$). The algorithm LEAFRECON is then run on those new reference leaves, and so on until the entire tree is recovered. (See Figure 6.) The algorithm is given in Figure 4. Some steps are detailed in Section 3. We denote estimates with hats, for example, the estimate of $P^{ar}$ is $\hat{P}^{ar}$.

*Subroutine* SEPRECON. For its part, the algorithm SEPRECON consists in taking a separator edge $(w, w')$ along with the leaf $a'$ from which it was found and the new reference leaf $a$ it led to, and computing

$$\hat{P}^{ww'} := (\hat{P}^{aw})^{-1}\hat{P}^{aa'}(\hat{P}^{w'a'})^{-1},$$

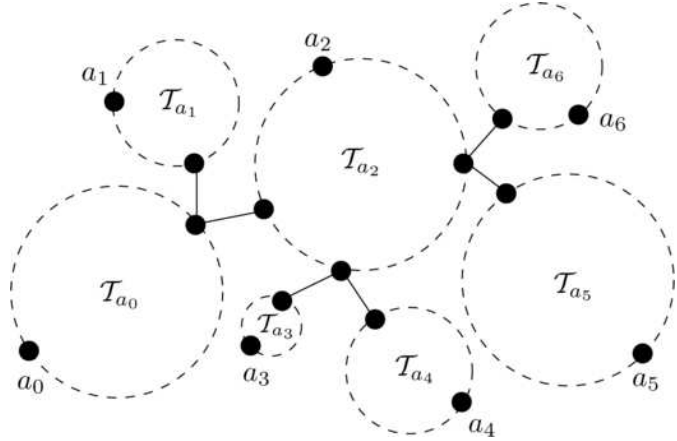

FIG. 6. *Schematic representation of the execution of* LEAFRECON. *The only edges shown are separators.*

where the matrices $\hat{P}^{aw}$, $\hat{P}^{w'a'}$ have been computed in the subtree reconstruction phase and $\hat{P}^{aa'}$ can be estimated from the data. It will be important in the error analysis that the two leaves $a, a'$ are at distance at most $\Delta$ from $w, w'$ respectively. We then use Bayes' rule to compute $\hat{P}^{w'w}$. See Figure 5.

2.3. *Modifications.* The previous description of the reconstruction algorithm is rather informal. Also, we are led to make a few modifications to the basic algorithm. Those are described where needed in the course of the analysis in the next section. Here is a list of the changes, all of which appear in the figures where the corresponding routines are detailed.

1. In Chang's spectral technique, it is crucial that the eigenvalues in ($\star$) be different and actually well separated. Below, we multiply the system ($\star$) by random Gaussians and obtain the new system ($\star'$). See "Separation of exact eigenvalues."
2. In ($\star'$), once the eigenvectors are recovered, they have to be normalized properly to obtain the estimated transition matrix $\tilde{P}^{ar}$. This is detailed in the subsection "Error on estimated eigenvectors."
3. All estimated transition matrices have to be made stochastic. This is done in subsection "Stochasticity."

**3. Error analysis.** As pointed out in the previous section, the distribution on the leaves is known only approximately through sampling. The purpose of this section is to account for the error introduced by this approximation.

For $\mathcal{W}$ a subset of vertices of $\mathcal{T}$, recall that $\pi_{\mathcal{W}}$ is the joint distribution on $\mathcal{W}$. We denote by $\hat{\pi}_{\mathcal{W}}$ our estimate of $\pi_{\mathcal{W}}$ obtained by using the estimated mutation matrices. For a vertex $u$, we let $\pi_u(\cdot) = \mathbb{P}[s(u) = \cdot]$. We denote by $\mathbb{1}$ the all-one vector (the size is usually clear from the context). For any vector $\rho$, we let $\text{diag}(\rho)$ be the diagonal matrix with diagonal $\rho$. Recall that for a vector $x$, $\|x\|_1 = \sum_i |x_i|$, and for a matrix $X$, $\|X\|_1 = \max_j \sum_i |x_{ij}|$. Recall that $[n]$ stands for the set of leaves. From now on, we assume that the tree is known and that the model is $(\beta, 0, n^{-\kappa_\pi})$-nonsingular, for $\beta, \kappa_\pi > 0$ constant. Theorems 1 and 2 both follow from the following theorem.

THEOREM 5. *Assume the tree is known and the model is $(\beta, 0, n^{-\kappa_\pi})$-nonsingular, for $\beta, \kappa_\pi > 0$ constant. For all $\varepsilon, \delta > 0$ and $n$ large enough, the reconstruction algorithm produces a Markov model satisfying*

$$\|\hat{\pi}_{[n]} - \pi_{[n]}\|_1 \leq \varepsilon,$$

*with probability at least $1 - \delta$. The running time of the algorithm is polynomial in $n, 1/\varepsilon, 1/\delta$.*

We can now prove Theorem 1.

PROOF OF THEOREM 1. First apply Theorem 3 to recover the topology. Then apply Theorem 5 to infer the transition matrices. □

The proof of Theorem 2 is similar—it uses Theorem 4 instead of Theorem 3. The proof is given in the Appendix.

Below we use the expression *with high probability* (w.h.p.) to mean with probability at least $1 - 1/\text{poly}(n)$. Likewise, we say *negligible* to mean at most $1/\text{poly}(n)$. In both definitions it is implied that $\text{poly}(n)$ is $O(n^K)$ for a constant $K$ that can be made as large as one wants if the number of samples is $O(n^{K'})$ with $K'$ large enough. Standard linear algebra results used throughout the analysis can be found, for example, in [20].

In the rest of this section, $\beta$, $\kappa_\pi$ and $k = |\mathcal{C}|$ are fixed constants. In particular, polynomial factors in $\beta$ and $k$ are dominated by polynomials in $n$.

3.1. *Approximate spectral argument.* In this subsection we address several issues arising from the application of Chang's spectral technique to an approximate distribution on the leaves. Our discussion is summarized in Proposition 1. We use the notation of Section 2.1.

PROPOSITION 1. *Let $a$ be a leaf and let $r$ be an internal vertex at distance at most $\Delta$ from $a$. Then there exists a relabeling of the states at $r$ so that the estimate $\hat{P}^{ar}$ recovered from ($\star$) using $\text{poly}(n)$ samples is such that the error $\|\hat{P}^{ar} - P^{ar}\|_1$ is negligible w.h.p.*

The relabeling issue mentioned in Proposition 1 will be tackled in Section 3.2.

*Determinants on paths.* The estimation error depends on the determinant of the transition matrices involved. Since we use Chang's spectral technique where $a \to r$, $r \to b$ and $r \to c$ are paths rather than edges, we need a lower bound on transition matrices over paths. This is where the use of short paths is important. Multiplicativity of determinants gives immediately that all determinants of transition matrices on paths of length $O(\Delta)$ are at least $1/\text{poly}(n)$.

LEMMA 1 (Bound on depth). *The depth $\Delta$ of any full binary tree is bounded above by $\log_2 n + 1$.*

PROOF. Because the tree is full, the inequality $\Delta \geq d$ implies that there is an edge on one side of which there is a complete binary subtree of depth $d$. Since there are only $n$ vertices in the tree, we must have $\Delta \leq \log_2 n + 1$. □

LEMMA 2 (Determinants on paths). *Fix $\theta > -2\log_2 \beta$ constant. Let $a, b$ be vertices at distance at most $2\Delta + 1$ from each other. Then, under the $(\beta, 0, n^{-\kappa_\pi})$-nonsingularity assumption, the transition matrix between $a$ and $b$ satisfies $|\det[P^{ab}]| \geq n^{-\theta}$ for $n$ large enough.*

PROOF. This follows from the observation that $P^{ab}$ is the product of the mutation matrices on the path between $a$ and $b$. Every matrix has its determinant at least $\beta$ (in absolute value). By the multiplicativity of determinants and Lemma 1,

$$|\det[P^{ab}]| \geq \beta^{2\Delta+1} \geq \beta^{2\log_2 n + 3} > n^{-\theta},$$

for $n$ large enough. □

*Error on leaf distributions.* The algorithm estimates leaf distributions through sampling. We need to bound the error introduced by sampling. Let $a, b, c$ be leaves at distance at most $2\Delta + 1$ from each other and consider the eigenvalue decomposition $(\star)$. We estimate $P^{ab}$ by taking $\text{poly}(n)$ samples and computing

$$\hat{P}^{ab}_{ij} = \frac{N^{a,b}_{i,j}}{N^a_i},$$

for $i, j \in \mathcal{C}$, where $N^a_i$ is the number of occurrences of $s(a) = i$ and $N^{a,b}_{i,j}$ is the number of occurrences of $s(a) = i, s(b) = j$. Likewise, for $P^{ab,\gamma}_{ij}$, we use $\text{poly}(n)$ samples and compute the estimate

$$\hat{P}^{ab,\gamma}_{ij} = \frac{N^{a,b,c}_{i,j,\gamma}}{N^a_i},$$

where $N^{a,b,c}_{i,j,\gamma}$ is the number of occurrences of $s(a) = i, s(b) = j, s(c) = \gamma$. We also bound the error on the 1-leaf distributions; this will be used in the next subsection. We use $\text{poly}(n)$ samples to estimate $\pi_a$ using empirical frequencies. Standard concentration inequalities give that $\|P^{ab} - \hat{P}^{ab}\|_1$, $\|P^{ab,\gamma} - \hat{P}^{ab,\gamma}\|_1$, and $\|\pi_a - \hat{\pi}_a\|_1$ are negligible w.h.p.

LEMMA 3 (Sampling error). *For all $e, p > 0$, there is an $s > 0$ such that if the number of samples is greater than $n^s$, then, with probability at least $1 - 1/n^p$, the estimation error on the matrices $P^{ab}$ and $P^{ab,\gamma}$ satisfies*

$$\|\hat{P}^{ab} - P^{ab}\|_1 \leq \frac{1}{n^e}, \qquad \|\hat{P}^{ab,\gamma} - P^{ab,\gamma}\|_1 \leq \frac{1}{n^e},$$

*for all $a, b \in \mathcal{L}$ and $\gamma \in \mathcal{C}$, and the estimation error on the leaf distributions satisfies*

$$\|\hat{\pi}_a - \pi_a\|_1 \leq \frac{1}{n^e},$$

*for all $a \in \mathcal{L}$, provided $n$ is large enough.*

PROOF. Note that the nonsingularity assumption ensures that, for each leaf $a$ and state $i$, if one uses a sample size $n^s$ with $s$ large enough, then w.h.p. there will be poly$(n)$ samples where $s(a) = i$. The bounds then follow from Hoeffding's inequality. □

*Separation of exact eigenvalues.* In Section 2 it was noted that the eigenvalues in $(\star)$ need to be distinct to guarantee that all eigenspaces have dimension 1. This is clearly necessary to recover the columns of the transition matrix $P^{ar}$. When taking into account the error introduced by sampling, we actually need more. From standard results on eigenvector sensitivity, it follows that we want the eigenvalues to be well separated. A polynomially small separation will be enough for our purposes. We accomplish this by using a variant of an idea of Chang [5] which consists in multiplying the matrix $P^{rc}$ in $(\star)$ by a random Gaussian vector. One can think of this as adding an extra edge $(c, d)$ and using leaves $a, b, d$ for the reconstruction, except that we do not need the transition matrix $P^{cd}$ to be stochastic and only one row of it suffices. More precisely, let $U$ be a vector whose $k$ entries are independent Gaussians with mean 0 and variance 1. We solve the eigenvalue problem $(\star)$ with $P^{rc}_{\cdot \gamma}$ replaced by $\Upsilon = (v_i)_{i=1}^k = P^{rc}U$, that is, we solve

$$(\star') \qquad P^{ab,U}(P^{ab})^{-1} = P^{ar} \operatorname{diag}(\Upsilon)(P^{ar})^{-1},$$

where $P^{ab,U}$ is a matrix whose $(i, j)$th entry is

$$\sum_{\gamma \in \mathcal{C}} U_\gamma \mathbb{P}[s(c) = \gamma, s(b) = j | s(a) = i].$$

To see that $(\star')$ holds, multiply the following equation (in matrix form) to the right by $(P^{ab})^{-1} = (P^{rb})^{-1}(P^{ar})^{-1}$:

$$[P^{ar} \operatorname{diag}(\Upsilon) P^{rb}]_{ij} = \sum_{h \in \mathcal{C}} P^{ar}_{ih} v_h P^{rb}_{hj}$$

$$= \sum_{h \in \mathcal{C}} P^{ar}_{ih} \left( \sum_{\gamma \in \mathcal{C}} P^{rc}_{h\gamma} U_\gamma \right) P^{rb}_{hj}$$

$$= \sum_{\gamma \in \mathcal{C}} U_\gamma \sum_{h \in \mathcal{C}} P^{ar}_{ih} P^{rc}_{h\gamma} P^{rb}_{hj}$$

$$= P^{ab,U}_{ij}.$$

A different (independent) vector $U$ is used for every triple of leaves considered by the algorithm. Next we show that w.h.p. the entries of $\Upsilon = (v_i)_{i=1}^k$ are $1/\operatorname{poly}(n)$-separated.

LEMMA 4 (Eigenvalue separation). *For all $d \geq \theta$ and $p \leq d - \theta$, with probability at least $1 - n^{-p}$, no two entries of $\Upsilon = (v_i)_{i=1}^k$ are at distance less than $n^{-d}$ for all $n$ large enough.*

PROOF. By Lemma 2, $|\det[P^{rc}]| \geq n^{-\theta}$. Take any two rows $i, j$ of $P^{rc}$. The matrix, say, $A$, whose entries are the same as $P^{rc}$ except that row $i$ is replaced by $P^{rc}_{i\cdot} - P^{rc}_{j\cdot}$, has the same determinant as $P^{rc}$. Moreover,

$$|\det[A]| \leq \sum_\sigma \prod_{h=1}^k |A_{h\sigma(h)}| \leq \prod_{h=1}^k \|A_{h\cdot}\|_1,$$

where the sum is over all permutations of $\{1, \ldots, k\}$. Therefore,

$$\|P^{rc}_{i\cdot} - P^{rc}_{j\cdot}\|_1 \geq n^{-\theta}.$$

By the Cauchy–Schwarz inequality,

$$\|P^{rc}_{i\cdot} - P^{rc}_{j\cdot}\|_2 \geq 1/(\sqrt{k} n^\theta).$$

Therefore, $(P^{rc}_{i\cdot} - P^{rc}_{j\cdot})U$ is Gaussian with mean 0 and variance at least $1/(k n^{2\theta})$. A simple bound on the normal distribution gives

$$\mathbb{P}\left[|(P^{rc}_{i\cdot} - P^{rc}_{j\cdot})U| < \frac{1}{n^d}\right] \leq 2 \frac{1}{n^d} \frac{\sqrt{k} n^\theta}{\sqrt{2\pi}}.$$

There are $O(k^2)$ pairs of rows to which we apply the previous inequality. The union bound gives the result. □

*Error on estimated l.h.s.* On the l.h.s. of $(\star')$, we use the following estimate $\hat{P}^{ab,U} = \sum_{\gamma \in \mathcal{C}} U_\gamma \hat{P}^{ab,\gamma}$. Below we show that the error on the l.h.s. of $(\star')$ is negligible w.h.p.

LEMMA 5 (Error on l.h.s.). *For all $e, p > 0$, there is an $s > 0$ such that if the number of samples is greater than $n^s$, then, with probability at least $1 - 1/n^p$, the error on the l.h.s. of $(\star')$ satisfies*

$$\|\hat{P}^{ab,U}(\hat{P}^{ab})^{-1} - P^{ab,U}(P^{ab})^{-1}\|_1 \leq \frac{1}{n^e}$$

*for all $n$ large enough.*

PROOF. From the submultiplicativity of $\|\cdot\|_1$, we obtain

$$\|\hat{P}^{ab,U}(\hat{P}^{ab})^{-1} - P^{ab,U}(P^{ab})^{-1}\|_1$$
(3)
$$\leq \|P^{ab,U}\|_1 \|(\hat{P}^{ab})^{-1} - (P^{ab})^{-1}\|_1 + \|(P^{ab})^{-1}\|_1 \|\hat{P}^{ab,U} - P^{ab,U}\|_1$$
$$+ \|(\hat{P}^{ab})^{-1} - (P^{ab})^{-1}\|_1 \|\hat{P}^{ab,U} - P^{ab,U}\|_1,$$

so it suffices to prove that each term on the r.h.s. can be made small enough.

First, note that, using a standard formula for the inverse, we have

$$(4) \qquad |(P^{ab})^{-1}_{ij}| = \frac{1}{|\det[P^{ab}]|} |(\mathrm{adj}[P^{ab}])_{ij}| \leq n^\theta,$$

where we have used the nonsingularity assumption, and the fact that the quantity $\mathrm{adj}[P^{ar}]_{ij}$ is the determinant of a substochastic matrix. Therefore, $\|(P^{ab})^{-1}\|_1 \leq kn^\theta$.

A standard linear algebra result [20] gives

$$\|(\hat{P}^{ab})^{-1} - (P^{ab})^{-1}\|_1 \leq \frac{\|(P^{ab})^{-1}\|_1 \|\hat{P}^{ab} - P^{ab}\|_1}{1 - \|(P^{ab})^{-1}\|_1 \|\hat{P}^{ab} - P^{ab}\|_1} \|(P^{ab})^{-1}\|_1$$

$$\leq \frac{2k^2 n^{2\theta}}{n^{e'}},$$

where $e'$ is the $e$ from Lemma 3 and is taken larger than $\theta$ so that the denominator on the first line is less than $1/2$.

We now compute the error on $P^{ab,U}$. We have

$$\|\hat{P}^{ab,U} - P^{ab,U}\|_1 \leq \sum_{\gamma \in \mathcal{C}} |U_\gamma| \|\hat{P}^{ab,\gamma} - P^{ab,\gamma}\|_1 \leq \frac{1}{n^{e'}} \|U\|_1.$$

Also,

$$\|P^{ab,U}\|_1 \leq \sum_{\gamma \in \mathcal{C}} |U_\gamma| \|P^{ab,\gamma}\|_1 \leq k\|U\|_1.$$

By a simple bound (see, e.g., [9]), for $g > 0$,

$$\mathbb{P}[|U_\gamma| \geq \sqrt{2\log(n^g)}] \leq 2\frac{1}{\sqrt{2\pi}} \frac{\exp(-2\log(n^g)/2)}{\sqrt{2\log(n^g)}} \leq \frac{1}{n^g \sqrt{\pi \log(n^g)}}.$$

So with probability at least $1 - k/(n^g \sqrt{\pi \log(n^g)})$, we get

$$\|U\|_1 \leq k\sqrt{2\log(n^g)}.$$

Taking $s$ (and therefore $e'$) large enough, the above bounds give

$$\|\hat{P}^{ab,U}(\hat{P}^{ab})^{-1} - P^{ab,U}(P^{ab})^{-1}\|_1$$
$$\leq \frac{2k^4 n^{2\theta} \sqrt{2\log(n^g)}}{n^{e'}} + \frac{k^2 n^\theta \sqrt{2\log(n^g)}}{n^{e'}} + \frac{2k^3 n^{2\theta} \sqrt{2\log(n^g)}}{n^{2e'}}$$
$$\leq \frac{1}{n^e}. \qquad \square$$

*Separation of estimated eigenvalues.* We need to make sure that the estimated l.h.s. of $(\star')$ is diagonalizable. By bounding the variation of the eigenvalues and relying on the gap between the exact eigenvalues, we show that the eigenvalues remain distinct and, therefore, $\hat{P}^{ab,U}(\hat{P}^{ab})^{-1}$ is diagonalizable.

LEMMA 6 (Sensitivity of eigenvalues). *For all $p > 0$, there is an $s > 0$ such that if the number of samples is greater than $n^s$, then, with probability at least $1 - 1/n^p$, the l.h.s. of $(\star')$, $\hat{P}^{ab,U}(\hat{P}^{ab})^{-1}$, is diagonalizable and all its eigenvalues are real and distinct. In particular, all eigenspaces have dimension 1.*

PROOF. Fix $d = p + \theta$ in Lemma 4. A standard theorem on eigenvalue sensitivity [20] asserts that if $\hat{v}_j$ is an eigenvalue of $\hat{P}^{ab,U}(\hat{P}^{ab})^{-1}$, there is an eigenvalue $v_i$ of $P^{ab,U}(P^{ab})^{-1}$ such that (recall that $P^{ar}$ is the matrix of eigenvectors)

$$
\begin{aligned}
|\hat{v}_j - v_i| &\leq \|P^{ar}\|_1 \|(P^{ar})^{-1}\|_1 \|\hat{P}^{ab,U}(\hat{P}^{ab})^{-1} - P^{ab,U}(P^{ab})^{-1}\|_1 \\
&\leq \frac{k^2 n^\theta}{n^e} \leq \frac{1}{3n^d},
\end{aligned}
\tag{5}
$$

where $e$ from Lemma 5 is taken large enough so that the last inequality holds. We have also used (4) from Lemma 5. Given that the separation between the entries of $\Upsilon$ is at least $1/n^d$ by Lemma 4, we deduce that there is a unique $v_i$ at distance at most $1/(3n^d)$ from $\hat{v}_j$ (note that $j$ might not be equal to $i$ since the ordering might differ in both vectors). This is true for all $j \in \mathcal{C}$. This implies that all $\hat{v}_j$'s are distinct and, therefore, they are real and $\hat{P}^{ab,U}(\hat{P}^{ab})^{-1}$ is diagonalizable as claimed. □

*Error on estimated eigenvectors.* From $(\star')$, we recover $k$ eigenvectors that are defined up to scaling. Assume that, for all $i \in \mathcal{C}$, $\hat{v}_i$ is the estimated eigenvalue corresponding to $v_i$ (see above). Denote by $\hat{X}^i$, $X^i$ their respective eigenvectors. We denote $\hat{X}$ (resp. $X$) the matrix formed with the $\hat{X}^i$'s (resp. $X^i$'s) as columns. Say we choose the estimated eigenvectors such that $\|\hat{X}^i\|_1 = 1$. This is not exactly what we are after because we need the rows to sum to 1 (not the columns). To fix this, we then compute $\eta = \hat{X}^{-1} \mathbb{1}$. This can be done because the columns of $\hat{X}$ form a basis. Then we define $\tilde{X}^i = \eta_i \hat{X}^i$ for all $i$ with the corresponding matrix $\tilde{X}$. Our final estimate $\tilde{P}^{ar} = \tilde{X}$ is a rescaled version of $\hat{X}$ with row sums 1. The careful reader may have noticed that some entries of $\tilde{X}$ may be negative. This is not an issue at this point. We will make sure in Lemma 12 that (one-step) mutation matrices are stochastic. Next we show that $\|\tilde{X} - X\|_1$ is negligible w.h.p.

LEMMA 7 (Sensitivity of eigenvectors). *For all $e, p > 0$, there is an $s > 0$ such that if the number of samples is greater than $n^s$, then, with probability at least $1 - 1/n^p$, we have*

$$\|\tilde{X} - X\|_1 \leq \frac{1}{n^e},$$

*for all $n$ large enough.*

PROOF. We want to bound the norm of $\tilde{X}^i - X^i$. We first argue about the components of $\hat{X}^i - X^i$ in the directions $X^j$, $j \neq i$. We follow a standard proof that can be found, for instance, in [17]. We need a more precise result than the one stated in the previous reference and so give the complete proof here.

Because the $X^i$'s form a basis, we can write

$$\hat{X}^i - X^i = \sum_{j \in \mathcal{C}} \rho_{ij} X^j,$$

for some values of $\rho_{ij}$'s. Denote $A = P^{ab,U}(P^{ab})^{-1}$, $\Delta_i = \hat{v}_i - v_i$ and $E = \hat{P}^{ab,U}(\hat{P}^{ab})^{-1} - P^{ab,U}(P^{ab})^{-1}$. Then

$$(A + E)\hat{X}^i = \hat{v}_i \hat{X}^i,$$

which, using $AX^i = v_i X^i$, implies

$$\sum_{j \in \mathcal{C}} v_j \rho_{ij} X^j + E\hat{X}^i = v_i \sum_{j \in \mathcal{C}} \rho_{ij} X^j + \Delta_i \hat{X}^i.$$

For all $j \in \mathcal{C}$, let $Z^j$ be the left eigenvector corresponding to $v_j$. It is well known that $(X^j)^T Z^{j'} = 0$ for all $j \neq j'$ (see, e.g., [17]). Fix $h \neq i \in \mathcal{C}$. Multiplying both sides of the previous display by $Z^h$ and rearranging gives

$$\rho_{ih} = \frac{(Z^h)^T(E\hat{X}^i) + \Delta_i(Z^h)^T \hat{X}^i}{(v_h - v_i)(Z^h)^T X^h}.$$

Here we make the $X^i$ be equal to the columns of $P^{ar}$ and the $Z^i$'s equal to the columns of $((P^{ar})^T)^{-1}$. In particular, we have $(Z^h)^T X^h = 1$. Recall that the $\hat{X}^i$'s were chosen such that $\|\hat{X}^i\|_1 = 1$. Fix $d = p + \theta$ in Lemma 4 so that $|v_h - v_i| \geq n^{-d}$. Choose the value of $e$ in Lemma 5 large enough so that the error $|\hat{v}_i - v_i|$ in (5) (ref. proof of Lemma 6 where $j$ is now $i$ by the construction above) is less than $1/n^{d+e'}$ for some fixed $e' > 0$ (the $d$ in Lemma 6 is $d + e'$ here). Then using standard matrix norm inequalities, the Cauchy–Schwarz inequality and Lemmas 4, 5 and 6, we get

$$|\rho_{ih}| = \left|\frac{(Z^h)^T(E\hat{X}^i) + \Delta_i(Z^h)^T \hat{X}^i}{(v_h - v_i)(Z^h)^T X^h}\right|$$

$$\leq \frac{\|Z^h\|_2 \|E\|_2 \|\hat{X}^i\|_2 + (1/n^{d+e'})\|Z^h\|_2 \|\hat{X}^i\|_2}{1/n^d}$$

$$\leq n^d \|Z^h\|_1 \|\hat{X}^i\|_1 [\sqrt{k} \|E\|_1 + 1/n^{d+e'}]$$

$$\leq \frac{2kn^\theta}{n^{e'}} \leq \frac{1}{n^{e''}},$$

for some $e'' > 0$, where we have used the bound $\|Z^h\|_1 \leq kn^\theta$ which follows from (4) in Lemma 6 and the choice of $Z^h$. We also used Lemma 5 to bound $\|E\|_1 \leq 1/(\sqrt{k} n^{d+e'})$ (which is possible if $s$ is large enough; remember that $k$ is a constant).

We now proceed to renormalize $\hat{X}$ appropriately. Define $\bar{X}^i = \hat{X}^i/(1+\rho_{ii})$. From the inequality above, we get

$$1 = \|\hat{X}^i\|_1 \leq |1 + \rho_{ii}| \|X^i\|_1 + \sum_{j \neq i} |\rho_{ij}| \|X^j\|_1$$

$$\leq k|1 + \rho_{ii}| + k^2/n^{e''}.$$

Assuming that $n^{e''}$ is large enough (i.e., choosing $e'$ above large enough), we get

$$|1 + \rho_{ii}| \geq \frac{1 - k^2/n^{e''}}{k} > 0.$$

Plugging $\bar{X}_i$ into the expansion of $\hat{X}_i$, we get

$$\|\bar{X}^i - X^i\|_1 = \left\| \sum_{j \neq i} \frac{\rho_{ij}}{1 + \rho_{ii}} X^j \right\|_1 \leq \frac{k^2}{n^{e''} - k^2} \leq \frac{1}{n^{e'''}},$$

for some $e''' > 0$, where we have used $\|X^j\|_1 \leq k$, $j \neq i$.

Denote $\bar{q} = \bar{X}\mathbb{1}$ the row sums of $\bar{X}$, the matrix formed with the $\bar{X}^i$'s as columns. The scaling between $\bar{X}$ and $\tilde{X}$ is given by $\bar{\eta} = \bar{X}^{-1}\mathbb{1}$. (Recall the definition of $\tilde{X}$ from the paragraph above the statement of the lemma.) Indeed, because the columns of $\tilde{X}$ and $\bar{X}$ are the same up to scaling, there is a vector $\tilde{\eta}$ such that $\tilde{X} = \bar{X} \operatorname{diag}(\tilde{\eta})$. By the normalization of both matrices, we get

$$\bar{X}\bar{\eta} = \tilde{X}\mathbb{1} = \bar{X}\operatorname{diag}(\tilde{\eta})\mathbb{1} = \bar{X}\tilde{\eta}.$$

Because $\bar{X}$ is invertible, $\tilde{\eta} = \bar{\eta}$. We want to argue that $\bar{\eta}$ is close to $\mathbb{1}$, that is, that $\bar{X}$ and $\tilde{X}$ are close. Note that

$$\|\bar{\eta} - \mathbb{1}\|_1 = \|\bar{X}^{-1}(\mathbb{1} - \bar{q})\|_1 \leq \|\bar{X}^{-1}\|_1 \|\mathbb{1} - \bar{q}\|_1.$$

By the condition $\|\bar{X}^i - X^i\|_1 \leq 1/n^{e'''}$ for all $i$ and the fact that the row sums of $X$ are 1, we get $\|\mathbb{1} - \bar{q}\|_1 \leq k/n^{e'''}$. To bound $\|\bar{X}^{-1}\|_1$, let $\bar{E} = \bar{X} - X$ and note that, using a standard theorem on the sensitivity of the inverse [20],

$$\|\bar{X}^{-1}\|_1 \leq \|(X+\bar{E})^{-1} - X^{-1} + X^{-1}\|_1$$
$$\leq \|X^{-1}\|_1 \frac{\|X^{-1}\|_1 \|\bar{E}\|_1}{1 - \|X^{-1}\|_1 \|\bar{E}\|_1} + \|X^{-1}\|_1$$
$$\leq \frac{\|X^{-1}\|_1}{1 - \|X^{-1}\|_1 \|\bar{E}\|_1}.$$

As we have seen before, $\|X^{-1}\|_1 \leq kn^\theta$ and by the bound above, $\|\bar{E}\|_1 \leq 1/n^{e'''}$. Assuming that $kn^\theta/n^{e'''} \leq 1/2$, we get

$$\|\bar{X}^{-1}\|_1 \leq 2\|X^{-1}\|_1 \leq 2kn^\theta.$$

Therefore,

$$\|\bar{\eta} - \mathbb{1}\|_1 \leq \frac{2k^2 n^\theta}{n^{e'''}}.$$

This finally gives the bound

$$\|\tilde{X} - X\|_1 \leq \|\tilde{X} - \bar{X}\|_1 + \|\bar{X} - X\|_1$$
$$\leq \|\bar{X}\operatorname{diag}(\bar{\eta}) - \bar{X}\|_1 + \frac{1}{n^{e'''}}$$
$$\leq \|\bar{X}\|_1 \|\bar{\eta} - \mathbb{1}\|_1 + \frac{1}{n^{e'''}}$$
$$\leq [\|\bar{X} - X\|_1 + \|X\|_1] \frac{2k^2 n^\theta}{n^{e'''}} + \frac{1}{n^{e'''}}$$
$$\leq \left[\frac{1}{n^{e'''}} + k\right] \frac{2k^2 n^\theta}{n^{e'''}} + \frac{1}{n^{e'''}} \leq \frac{1}{n^e},$$

if $e'''$ (i.e., $e'$) is large enough (where $e$ on the last line is the one in the statement of Lemma 7). □

There is one last issue which is that $X$ is the same as $P^{ar}$ *up to permutation on the states of* $r$. But since relabeling internal nodes does not affect the output distribution, we assume w.l.o.g. that $P^{ar} = X$. We make sure in the next subsection that this relabeling is performed only once for each internal node.

3.2. *Bounding error propagation.* The correctness of the algorithm proceeds from the following remarks.

*Partition.* We have to check that the successive application of LEAFRE-CON covers the entire tree, that is, that all edges are reconstructed. Figure 6 helps in understanding why this is so. When uncovering a separator edge, we associate to it a new reference leaf at distance at most $\Delta$. This can always be done by definition of $\Delta$. It also guarantees that the subtree associated to this new leaf will cover the endpoint of the separator outside the subtree from which it originated. This makes the union of all subtrees explored at any point in the execution (together with their separators) connected. It follows easily that the entire tree is eventually covered.

LEMMA 8 (Partition). *The successive application of* LEAFRECON *covers the entire tree.*

PROOF. We need to check that the algorithm outputs a transition matrix for each edge in $\mathcal{T}$. Denote $\mathcal{T}_{a_t}$ the subtree explored by LEAFRECON applied to $a_t$. The key point is that, for all $t$, the tree $\mathcal{T}_{\leq t}$ made of all $\mathcal{T}_{a_{t'}}$ for $t' \leq t$, as well as their separators, is connected. We argue by induction. This is clear for $t = 0$. Assume this is true for $t$. Because $\mathcal{T}$ is a tree, $\mathcal{T}_{\leq t}$ is a (connected) subtree of $\mathcal{T}$ and $(w_{t+1}, w'_{t+1})$ is an edge on the "boundary" of $\mathcal{T}_{\leq t}$, the leaf $a_{t+1}$ lies outside $\mathcal{T}_{\leq t}$. Moreover, being chosen as the closest leaf from $w'_{t+1}$, it is at distance at most $\Delta$. Therefore, applying LEAFRECON to $a_{t+1}$ will cover a (connected) subtree *including* $w'_{t+1}$. This proves the claim. □

*Subroutines.* Using Lemma 7 and standard linear algebraic inequalities, we show that the (unnormalized) estimates computed in LEAFRECON and SEPRECON have negligible error w.h.p.

LEMMA 9 (Error analysis: LEAFRECON). *Let $a$ be a leaf. For all $e, p > 0$, there is an $s > 0$ such that if the number of samples is greater than $n^s$, then, with probability at least $1 - 1/n^p$, all edges $(r_0, r)$ reconstructed by* LEAFRECON *applied to $a$ satisfy*

$$\|\tilde{P}^{r_0 r} - P^{r_0 r}\|_1 \leq \frac{1}{n^e},$$

(*after a proper relabeling of the rows and columns of $P^{r_0 r}$ to match the labeling of $\tilde{P}^{r_0 r}$*), *and also*

$$\|\tilde{\pi}_r - \pi_r\|_1 \leq \frac{1}{n^e},$$

(*after a proper relabeling of the vertices*) *for all $n$ large enough.*

PROOF. Note that

$$\|\tilde{P}^{r_0 r} - P^{r_0 r}\|_1 = \|(\tilde{P}^{a r_0})^{-1} \tilde{P}^{ar} - (P^{a r_0})^{-1} P^{ar}\|_1,$$

and, thus, a calculation identical to the proof of Lemma 5 shows that the above error can be made negligible w.h.p. Also,

$$\|\tilde{\pi}_r - \pi_r\|_1 = \|\hat{\pi}_a \tilde{P}^{ar} - \pi_a P^{ar}\|_1$$
$$\leq \|\hat{\pi}_a\|_1 \|\tilde{P}^{ar} - P^{ar}\|_1 + \|\pi_a - \hat{\pi}_a\|_1 \|P^{ar}\|_1,$$

so, by Lemmas 3 and 5, $\|\tilde{\pi}_r - \pi_r\|_1$ can be made negligible w.h.p.

The algorithm computes the estimate $\tilde{P}^{rr_0}$ by Bayes' rule. Therefore,

$$|\tilde{P}^{rr_0}_{ij} - P^{rr_0}_{ij}| = \left|\frac{\tilde{\pi}_{r_0}(j)\tilde{P}^{r_0r}_{ji}}{\tilde{\pi}_r(i)} - \frac{\pi_{r_0}(j)P^{r_0r}_{ji}}{\pi_r(i)}\right|$$
$$\leq \frac{\tilde{\pi}_{r_0}(j)}{\tilde{\pi}_r(i)}|\tilde{P}^{r_0r}_{ji} - P^{r_0r}_{ji}| + \left|\frac{\tilde{\pi}_{r_0}(j)}{\tilde{\pi}_r(i)} - \frac{\pi_{r_0}(j)}{\pi_r(i)}\right| P^{r_0r}_{ji}.$$

Assume $\|\tilde{P}^{r_0r} - P^{r_0r}\|_1 \leq n^{-e'}$ and $\|\tilde{\pi}_r - \pi_r\|_1 \leq n^{-e''}$ with $e'' > \kappa_\pi$. Then,

$$\left|\frac{\tilde{\pi}_{r_0}(j)}{\tilde{\pi}_r(i)} - \frac{\pi_{r_0}(j)}{\pi_r(i)}\right| \leq \frac{\pi_r(i)|\tilde{\pi}_{r_0}(j) - \pi_{r_0}(j)| + \pi_{r_0}(j)|\tilde{\pi}_r(i) - \pi_r(i)|}{\tilde{\pi}_r(i)\pi_r(i)}$$
$$\leq \frac{2n^{-e''}}{(n^{-\kappa_\pi} - n^{e''})^2}.$$

Therefore,

$$|\tilde{P}^{rr_0}_{ij} - P^{rr_0}_{ij}| \leq \frac{(1 + n^{-e''})n^{-e'}}{n^{-\kappa_\pi} - n^{-e''}} + \frac{2n^{-e''}}{(n^{-\kappa_\pi} - n^{e''})^2}.$$

The r.h.s. can be made negligible with a large enough sample size (i.e., large enough $e', e''$).  □

LEMMA 10 (Error analysis: SEPRECON).  *For all $e, p > 0$, there is an $s > 0$ such that if the number of samples is greater than $n^s$, then, with probability at least $1 - 1/n^p$, every edge $(w, w')$ reconstructed by SEPRECON satisfies*

$$\|\tilde{P}^{ww'} - P^{ww'}\|_1 \leq \frac{1}{n^e},$$

*(after permuting the rows and columns of $P^{ww'}$ to match the labeling of $\tilde{P}^{ww'}$) for all $n$ large enough.*

PROOF.  Let $a, a'$ be the leaves used by SEPRECON to estimate $P^{ww'}$. Then,

$$\|\tilde{P}^{ww'} - P^{ww'}\|_1 = \|(\tilde{P}^{aw})^{-1}\hat{P}^{aa'}(\tilde{P}^{w'a'})^{-1} - (P^{aw})^{-1}P^{aa'}(P^{w'a'})^{-1}\|_1.$$

By applying twice an inequality of the form (3), the r.h.s. can be bounded above by a sum of terms involving primarily $\|\tilde{P}^{aw} - P^{aw}\|_1$, $\|\tilde{P}^{a'w'} - P^{a'w'}\|_1$, and $\|\hat{P}^{aa'} - P^{aa'}\|_1$. Those errors can be made negligible w.h.p. by Lemmas 3 and 5. The algorithm then uses Bayes' rule to compute $\tilde{P}^{w'w}$. The error on that estimate can be obtained by a calculation identical to that in Lemma 9. □

*Consistency.* Next we prove that all choices of labelings are done consistently. This follows from the fact that, for each node, say, $w$, the arbitrary labeling is performed only once. Afterward, all computations involving $w$ use only the matrix $\tilde{P}^{aw}$, where $a$ is the reference leaf for $w$.

LEMMA 11 (Consistency). *The labelings are made consistently by subroutines* LEAFRECON *and* SEPRECON.

PROOF. We briefly sketch the proof. By Lemma 7, we know that, for a reference leaf $a$ and an internal node $r$, the estimated transition matrix $\tilde{P}^{ar}$ is close to the exact transition matrix $P^{ar}$ *after properly relabeling the columns of $P^{ar}$ to match the arbitrary labeling of the columns of $\tilde{P}^{ar}$.* Let $\Gamma_r$ be the permutation matrix performing this relabeling on the columns of $P^{ar}$, that is, such that $\|\tilde{P}^{ar} - P^{ar}\Gamma_r\|_1$ is small. Let $\Gamma_{r_0}$ be the corresponding matrix for node $r_0$. Then, the matrix $P^{r_0r}$ (which, contrary to Lemma 9, we assume *not* to have been relabeled according to $\tilde{P}^{r_0r}$) satisfies the equation

$$(P^{ar_0}\Gamma_{r_0})^{-1}P^{ar}\Gamma_r = \Gamma_{r_0}^{-1}(P^{ar_0})^{-1}P^{ar}\Gamma_r$$
$$= \Gamma_{r_0}^T P^{r_0r}\Gamma_r.$$

The last line is the matrix $P^{r_0r}$ after being properly relabeled according the arbitrary choices made by LEAFRECON at nodes $r_0, r$. By Lemmas 7 and 9, this implies that $\|\tilde{P}^{r_0r} - \Gamma_{r_0}^T P^{r_0r}\Gamma_r\|_1$ is small as required. A similar argument applies to the computation of $\tilde{\pi}_r$, $\tilde{\pi}_{r_0}$ and $\tilde{P}^{rr_0}$ in LEAFRECON, as well as the computation of $\tilde{P}^{ww'}$, $\tilde{P}^{w'w}$, $\tilde{\pi}_w$ and $\tilde{\pi}_{w'}$ in SEPRECON. □

*Stochasticity.* It only remains to make the estimates of mutation matrices into stochastic matrices. Say $\tilde{P}^{ww'}$ is the (unnormalized) estimate of $P^{ww'}$. First, some entries might be negative. Define $\tilde{P}_+^{ww'}$ to be the positive part of $\tilde{P}^{ww'}$. Then renormalize to get our final estimate

$$(6) \qquad \hat{P}_{i\cdot}^{ww'} = \frac{(\tilde{P}_+^{ww'})_{i\cdot}}{\|(\tilde{P}_+^{ww'})_{i\cdot}\|_1},$$

for all $i \in \mathcal{C}$. We know from Lemmas 9 and 10 that $\tilde{P}^{ww'}$ is close to $P^{ww'}$ in $L_1$ distance. From this, we show below that $\tilde{P}_+^{ww'}$ is also close to $P^{ww'}$ and $\|(\tilde{P}_+^{ww'})_{i\cdot}\|_1$ is close to 1 and that therefore $\|\hat{P}^{ww'} - P^{ww'}\|_1$ is negligible w.h.p.

LEMMA 12 (Stochasticity). *For all $e, p > 0$, there is an $s > 0$ such that if the number of samples is greater than $n^s$, then, with probability at least $1 - 1/n^p$, the estimate $\hat{P}^{ww'}$ is well defined and satisfies*

$$\|\hat{P}^{ww'} - P^{ww'}\|_1 \leq \frac{1}{n^e},$$

(*after permuting the rows and columns of $P^{ww'}$ to match the labeling of $\hat{P}^{ww'}$*) *for all $n$ large enough.*

PROOF. Because $P^{ww'}$ is nonnegative, taking the positive part of $\tilde{P}^{ww'}$ can only make it closer to $P^{ww'}$, that is,

$$\|\tilde{P}_+^{ww'} - P^{ww'}\|_1 \leq \|\tilde{P}^{ww'} - P^{ww'}\|_1.$$

Assume that, by Lemmas 9 and 10, we have the bound $\|\tilde{P}^{ww'} - P^{ww'}\|_1 \leq n^{-e'}$. Then the row sums of $\tilde{P}^{ww'}$ are at least $1 - kn^{-e'}$. Also, taking the positive part of $\tilde{P}^{ww'}$ can only decrease its row sums by $kn^{-e'}$. Therefore, we get

$$\|(\tilde{P}_+^{ww'})_{i\cdot}\|_1 \geq 1 - \frac{2k}{n^{e'}}.$$

Thus,

$$\|\hat{P}^{ww'} - P^{ww'}\|_1 \leq \|\tilde{P}_+^{ww'} - P^{ww'}\|_1 + \|\hat{P}^{ww'} - \tilde{P}_+^{ww'}\|_1$$
$$\leq \|\tilde{P}^{ww'} - P^{ww'}\|_1 + \frac{2k}{n^{e'}}\|\hat{P}^{ww'}\|_1 \leq \frac{1 + 2k^2}{n^{e'}},$$

which can be made smaller than $n^{-e}$. $\square$

Note that we do not renormalize the node distributions because we only need to know the distribution at one arbitrary node and that node can conveniently be chosen among the leaves.

*Precision and confidence.* Now that all matrices have been approximately reconstructed, we prove that the distributions on the leaves of the estimated and real models are close. We show below that $\|\hat{\pi}_{[n]} - \pi_{[n]}\|_1$ is negligible w.h.p., thereby proving Theorem 5.

LEMMA 13 (Precision and confidence). *Let $\varepsilon, \delta > 0$. Using at most $\mathrm{poly}(n, 1/\varepsilon, 1/\delta)$ samples, with probability at least $1 - \delta$, the reconstructed model satisfies*

$$\|\hat{\pi}_{[n]} - \pi_{[n]}\|_1 \leq \varepsilon.$$

PROOF. We only give a quick sketch. Assume that the number of samples is taken large enough so that, by the sequence of lemmas above, the bounds on the $L_1$ error on the estimated transition matrices and on the estimated node distributions, $n^{-e}$, is smaller than $\varepsilon/(2nk)$ with probability at least $1 - \delta$. By the triangle inequality, $\|\hat{\pi}_{[n]} - \pi_{[n]}\|_1 \leq \|\hat{\pi}_\mathcal{V} - \pi_\mathcal{V}\|_1$, so it suffices to bound the $L_1$ error on the entire tree. Now, couple the exact model and the estimated model in a standard fashion. We seek to bound the probability that the two models differ at any vertex. Fix an arbitrary root. The probability that the models differ at the root is $\varepsilon/(2nk)$ by assumption. Stop if that happens. Otherwise, at each transition, the probability that the transition is different in the two models is less than $\varepsilon/(2n)$ (provided that they start from the same initial state). Again, if that happens, stop. Since there are at most $2n$ transitions, by the union bound, the probability that we stop at any step in the process is $\varepsilon$. $\square$

**4. Concluding remarks.** Many extensions of this work deserve further study:

- There remains a gap between our positive result (for general trees), where we require determinants $\Omega(1)$ and the hardness result which uses determinants exactly 0. Is learning possible when determinants are $\Omega(n^{-c})$ or even $\Omega(\log^{-c} n)$ (as it is in the case of HMMs)?
- There is another gap arising from the upper bound on the determinants. Having mutation matrices with determinant 1 does not seem like a major issue. It does not arise in the estimation of the mutation matrices. But it is tricky to analyze rigorously how the determinant 1 edges affect the reconstruction of the topology.
- We have emphasized the difference between $k=2$ and $k \geq 3$. As it stands, our algorithm works only for nonsingular models even when $k = 2$. It would be interesting to rederive the results of [7] using our technique.

## APPENDIX: PROOFS FOR THE CATERPILLAR CASE

We first sketch the proof for the topology reconstruction in the caterpillar case.

THEOREM 4. *Let $\zeta > 0, \kappa_\pi > 0$ and suppose that $\mathbf{M}$ consists of all matrices $P$ satisfying $n^{-\zeta} < |\det P| \leq 1$. Then for all $\theta > 0, \tau' > 0$, and all $T \in (\mathbf{TC}_3 \otimes \mathbf{M}, n^{-\kappa_\pi})$, one can recover from $n^{O(\zeta+\theta+\tau')}$ samples a topology $\mathcal{T}'$ with probability $1 - n^{-\theta}$, where the topology $\mathcal{T}'$ satisfies the following. It is obtained from the true topology $\mathcal{T}$ by contracting some of the internal edges whose corresponding mutation matrices $P$ satisfy $|\det P| > 1 - n^{-\tau'}$.*

PROOF (SKETCH). We use a distance-based method similar to [10, 11]. For general Markov models of evolution, Steel [31] introduced the following metric, known as log-det distance. Let $\mathcal{P}_{ab}$ be the set of edges on the (unique) path between leaves $a, b$. Define the matrix $F_{ab} = [f_{ab}(i,j)]_{i,j \in \mathcal{C}}$, where $f_{ab}(i,j) = \mathbb{P}[s(a) = i, s(b) = j]$. Then, Steel [31] showed that the quantity $\Psi_{ab} \equiv -\ln|\det(F_{ab})|$ defines a tree metric on the set of leaves by deriving the identity $\Psi_{ab} = \sum_{(u,v)\in\mathcal{P}_{ab}} \nu_{uv}$, where

$$\nu_{uv} = -\ln|\det(P^{uv})| - \tfrac{1}{2}\ln\left(\prod_{i\in\mathcal{C}}\pi_u(i)\right) + \tfrac{1}{2}\ln\left(\prod_{i\in\mathcal{C}}\pi_v(i)\right).$$

Below, we will need a slightly different expression. Noting that $P^{vu}$ is the "time-reversal" of $P^{uv}$, one immediately obtains

$$\nu_{uv} = -\tfrac{1}{2}\ln|\det(P^{uv})| - \tfrac{1}{2}\ln|\det(P^{vu})|.$$

A crucial observation in [10, 11] is that, to obtain good estimates of distances with a polynomial number of samples, one has to consider only pairs of leaves at a "short" distance. We note $\hat{\Psi}_{ab}$ the estimate of $\Psi_{ab}$. For $\Delta > 0$, define

$$S_\Delta = \{(a,b) \in \mathcal{L} \times \mathcal{L} : \hat{\Psi}_{ab} > 2\Delta\}.$$

Let $\Delta = -\ln[6n^{-\zeta}]$. Then it follows from [11], Proof of Theorem 14, that, for any $e, p > 0$, there exists an $s > 0$ large enough so that, using $n^s$ samples, with probability at least $1 - n^{-p}$, one has, for all $(a,b)$ in $S_{2\Delta}$,

$$|\hat{\Psi}_{ab} - \Psi_{ab}| < -\ln[1 - n^{-e}] \leq n^{-e},$$

and $S_{2\Delta}$ contains all pairs of leaves with $\Psi_{ab} \leq 2\Delta$, but no pair with $\Psi_{ab} > 6\Delta$.

We now consider quartets of leaves at a short distance. Define

$$Z_{2\Delta} = \{q \in \mathcal{L}^4 : \forall (a,b) \in q, (a,b) \in S_{2\Delta}\}.$$

We then use the four-point method to reconstruct quartets in $Z_{2\Delta}$: if $q$ is made of leaves $a, b, c, d$, then (w.l.o.g.) we infer the split $\{a,b\}\{c,d\}$, where

$$\hat{\Psi}_{ab} + \hat{\Psi}_{cd} \leq \min\{\hat{\Psi}_{ac} + \hat{\Psi}_{bd}, \hat{\Psi}_{ad} + \hat{\Psi}_{bc}\}.$$

By [10], Lemma 5, this is guaranteed to return the right topology if, for all $a', b' \in q$,

$$|\hat{\Psi}_{a'b'} - \Psi_{a'b'}| < \frac{x}{2},$$

where $x$ is the length (in the log-det distance) of the internal edge in the subtree induced by $q$. In other words, if $Q$ is the transition matrix on the internal edge of $q$, we can only reconstruct the topology of $q$ if $|\det(Q)|$ is bounded away from 1. Therefore, we define a threshold $\delta = -\ln[1 - n^{-\tau}]$ for some $\tau > 0$ and infer the topology only on those quartets in $S_{2\Delta}$ such that (w.l.o.g.)

$$\hat{\Psi}_{ab} + \hat{\Psi}_{cd} \leq \min\{\hat{\Psi}_{ac} + \hat{\Psi}_{bd}, \hat{\Psi}_{ad} + \hat{\Psi}_{bc}\} - 2\delta.$$

If we further impose the restriction $e > \tau$, then for $n$ large enough, all such quartets are correctly reconstructed w.h.p.

To contract small internal edges in a consistent fashion, we consider the following construction. Consider an undirected graph $H$ with node set $\mathcal{L}$. For two nodes $a, b$, there is an edge $(a, b)$ if $(a, b) \in S_{2\Delta}$ and if all reconstructed quartets put $a$ and $b$ on the same side of their corresponding split. Let $\{H_\alpha\}_{\alpha=1}^A$ be the connected components of $H$. Ultimately, our inferred tree (say, $\mathcal{T}'$) will be such that all leaves in a same connected component of $H$ form a star in $\mathcal{T}'$. To justify this approach, we make the following observations. Denote $u$ (resp. $v$) the internal node of $\mathcal{T}$ closest to $a$ (resp. $b$). If

$$-\tfrac{1}{2}\ln|\det(P^{uv})| \leq -\ln[1-n^{-\tau}] + \ln[1-n^{-e}],$$

then w.h.p. there exists an edge between $a$ and $b$ in $H$. Conversely, the existence of an edge $(a, b)$ in $H$ implies

$$-\tfrac{1}{2}\ln|\det(P^{uv})| \leq -\ln[1-n^{-\tau}] - \ln[1-n^{-e}],$$

or

$$|\det(P^{uv})| \geq [1-n^{-\tau}]^2[1-n^{-e}]^2.$$

Because there are at most $n$ nodes in each component of $H$, if $a$ and $b$ are in the same connected component, we have

$$|\det(P^{uv})| \geq 1 - n^{-\tau'},$$

where $\tau' < 2\tau - 1$. Also, let $a$ and $b$ be two nodes in $H$ connected by an edge and assume there is a leaf $c$ in $\mathcal{T}$ between $a$ and $b$ (i.e., $c$ is sticking out of the path between $u$ and $v$). Let $w$ be the internal node in $\mathcal{T}$ closest to $c$. Then we must have

$$-\tfrac{1}{2}\ln\min\{|\det(P^{uw})|, |\det(P^{wv})|\} \leq \tfrac{1}{2}[-\ln[1-n^{-\tau}] - \ln[1-n^{-e}]]$$
$$\leq -\ln[1-n^{-\tau}] + \ln[1-n^{-e}],$$

for $n$ large enough and, therefore, we are guaranteed to have either $(a, c)$ or $(b, c)$ in $H$ w.h.p. All these observations imply that each connected component of $H$ corresponds to a group of consecutive leaves with an internal path having a transition matrix close to identity. Therefore, we can assume that the connected components of $H$ form stars in $\mathcal{T}'$.

Finally, we choose a representative leaf $l_\alpha$ from each connected component $H_\alpha$. Let $\mathcal{T}''$ be the subtree of $\mathcal{T}$ induced by $l_1, \ldots, l_A$. It suffices to estimate $\mathcal{T}''$. Then the final estimate $\mathcal{T}'$ is $\mathcal{T}''$, where all representative leaves are replaced by their corresponding star.

The inference of $\mathcal{T}''$ is straightforward. Note first that if $l_\alpha$ and $l_\beta$ are leaves in $\mathcal{T}''$ and $u$ and $v$ are their respective closest internal nodes in $\mathcal{T}$, then

$$\nu_{uv} \geq -\ln[1-n^{-\tau}] + \ln[1-n^{-e}].$$

Also, by our choice of $\Delta$ and the construction of $H$, if $l_\alpha$ and $l_\beta$ are two consecutive leaves in $\mathcal{T}''$, then there is at least one reconstructed quartet where $l_\alpha$ and $l_\beta$ are on different sides of the split, that is, each edge in the tree $\mathcal{T}''$ is represented by a split in the reconstructed quartets. To construct $\mathcal{T}''$, we proceed by induction. We recall that a *cherry* in a tree is a pair of leaves whose topological distance is exactly 2. We first identify a cherry by finding a pair of leaves which is always on the same side of any split. This will be one of the two terminal cherries of $\mathcal{T}''$. Then we remove one of the two leaves in that cherry, and start over. This way, we build the tree $\mathcal{T}''$ a leaf at a time from one end to the other end. □

THEOREM 2. *Let $\phi_d, \kappa_\pi > 0$ be constants. Let $\mathcal{C}$ be a finite set and $\mathbf{M}$ denote the collection of $|\mathcal{C}| \times |\mathcal{C}|$ transition matrices $P$, where $1 \geq |\det P| > n^{-\phi_d}$. Then there exists a PAC-learning algorithm for $(\mathbf{TC}_3(n) \otimes \mathbf{M}_n, n^{-\kappa_\pi})$. The running time and sample complexity of the algorithm is $\operatorname{poly}(n, k, 1/\varepsilon, 1/\delta)$.*

PROOF (SKETCH). From Theorem 4, we can infer a tree $\mathcal{T}'$ which is obtained from the true topology $\mathcal{T}$ by contracting some of the internal edges whose corresponding mutation matrices $P$ satisfy $|\det P| > 1 - n^{-\tau'}$ (refer to the proof of Theorem 4 for notation). Now, note that the impossibility to infer (efficiently) quartets with a very small internal edge is of no consequence for the following reason. It is not hard to show that a stochastic matrix with a determinant close to 1 (in absolute value) is close to a permutation matrix. More precisely, for any $\tau' > 0$, there is an $e' > 0$ such that if $Q$ is a stochastic matrix with $|\det(Q)| \geq 1 - n^{-\tau'}$, then there is a permutation matrix $J$ such that $\|J - Q\|_1 \leq n^{-e'}$ (we omit the proof). Let $E_{\tau'}$ be the set of such transition matrices in our Markov model (only those corresponding to internal edges). By relabeling the states at the internal nodes, we can assume w.l.o.g. that all transition matrices on $E_{\tau'}$ are actually close to the identity matrix. Then if $e'$ is large enough, any realization of the Markov model is such that there is no transition on edges in $E_{\tau'}$ w.h.p. Put differently, from a PAC learning point of view, we can contract any edge in $E_{\tau'}$. □

**Acknowledgments.** We thank Jon Feldman, Ryan O'Donnell, Rocco Servedio and Mike Steel for interesting discussions.

DEPARTMENT OF STATISTICS
UNIVERSITY OF CALIFORNIA, BERKELEY
367 EVANS HALL
BERKELEY, CALIFORNIA 94720-3860
USA
E-MAIL: mossel@stat.berkeley.edu
       sroch@stat.berkeley.edu